%% file: acl_latex.tex
\pdfoutput=1

\documentclass[11pt]{article}
\usepackage{amsmath}
\usepackage{witharrows}

\usepackage[final]{acl}

\usepackage{times}
\usepackage{latexsym}

\usepackage[T1]{fontenc}

\usepackage[utf8]{inputenc}

\usepackage{microtype}

\usepackage{inconsolata}
\usepackage{enumitem}
\usepackage{graphicx}
\usepackage{bbold}
\usepackage{booktabs}
\usepackage{dsfont}
\usepackage{caption}
\usepackage{subcaption}
\usepackage{xspace}

\usepackage{syntonly}

\input{sections/math_commands}

\input{sections/notations}

\title{Successfully Guiding Humans with Imperfect Instructions by Highlighting Potential Errors and Suggesting Corrections}

\author{$^{\spadesuit}$Lingjun Zhao \and
  $^{\clubsuit}$Nguyen X. Khanh  \and $^{\spadesuit}$Hal Daum\'e III \\
  $^{\spadesuit}$University of Maryland, College Park \ $^{\clubsuit}$University of California, Berkeley 
   \\
  \texttt{lzhao123@umd.edu } \\}

\begin{document}
\maketitle
\begin{abstract}
\input{sections/abstract}

\end{abstract}

\section{Introduction} \label{introduction}
\input{sections/introduction}

\section{Related Work} \label{related_work}
\input{sections/related_work}

\section{Problem Setting} \label{problem}
\input{sections/problem}

\section{\ourmodel: Hallucination Detection and Remedy} \label{model}
\input{sections/model}

\section{Experiments} \label{experiments}
\input{sections/experiments}

\section{Conclusion} \label{conclusion}
\input{sections/conclusion}

\section*{Limitations}
\input{sections/limitations}

\section*{Acknowledgements}

We thank Hyemi Song, Yue Feng and Mingyang Xie for providing suggestions on improving human evaluation interface. We thank Eleftheria Briakou, Connor Baumler, Trista Cao, Navita Goyal and other group members for providing suggestions on human evaluation experimental design.

\bibliography{custom}

\clearpage
\onecolumn

\appendix

\section{Appendices}
\input{sections/appendix}

\end{document}

%% file: sections/math_commands.tex
\usepackage{amsmath,amsfonts,bm}

\def\eqref#1{equation~\ref{#1}}

\def\1{\bm{1}}

\DeclareMathAlphabet{\mathsfit}{\encodingdefault}{\sfdefault}{m}{sl}
\SetMathAlphabet{\mathsfit}{bold}{\encodingdefault}{\sfdefault}{bx}{n}

\newcommand{\sigmoid}{\sigma}



%% file: sections/notations.tex
\definecolor{lightgray}{HTML}{808080}
\definecolor{perfup}{HTML}{008000}
\definecolor{perfdown}{HTML}{1d7b21}
\definecolor{myred}{HTML}{e60035}

\newcommand\redsout{\bgroup\markoverwith{\textcolor{red}{\rule[0.5ex]{2pt}{2pt}}}\ULon}
\newcommand\bluesin{\bgroup\markoverwith{\textcolor{green}{\rule[-0.5ex]{2pt}{2pt}}}\ULon}
\newcommand\redsin{\bgroup\markoverwith{\textcolor{red}{\rule[-0.5ex]{2pt}{2pt}}}\ULon}

\renewcommand{\sectionautorefname}{\S\kern-0.2em}
\renewcommand{\subsectionautorefname}{\S\kern-0.2em}

\definecolor{lightgreen}{HTML}{1d7b21}

\renewcommand{\vec}[1]{\boldsymbol{#1}}

\newcommand{\instr}{\vec w}
\newcommand{\traj}{\vec r}

\newcommand{\correctphrase}{\hat{\vec w}}
\newcommand{\ourmodel}{HEAR\xspace}

%% file: sections/abstract.tex
Language models will inevitably err in situations with which they are unfamiliar. However, by effectively communicating uncertainties, they can still guide humans toward making sound decisions in those contexts. We demonstrate this idea by developing \ourmodel, a system that can successfully guide humans in simulated residential environments despite generating potentially inaccurate instructions. 
Diverging from systems that provide users with only the instructions they generate, \ourmodel warns users of potential errors in its instructions and suggests corrections. 
This rich uncertainty information effectively prevents misguidance and reduces the search space for users. 
Evaluation with 80 users shows that \ourmodel achieves a 13\% increase in success rate and a 29\% reduction in final location error distance compared to only presenting instructions to users. Interestingly, we find that offering users possibilities to explore, \ourmodel motivates them to make more attempts at the task, ultimately leading to a higher success rate. 
To our best knowledge, this work is the first to show the practical benefits of uncertainty communication in a long-horizon sequential decision-making problem.\footnote{Our code and data for model and human evaluation are publicly released at \hyperlink{https://lingjunzhao.github.io/HEAR.html}{https://lingjunzhao.github.io/HEAR.html}.} 
\looseness=-1

%% file: sections/introduction.tex
Expecting language models to consistently make accurate predictions in a dynamic world is unrealistic \cite{kalai2023calibrated,xu2024hallucination}.
Evidence shows that these models often falter in unfamiliar situations \cite{wu2023reasoning,dziri2024faith}.
Given the inherent fallibility of language models, an important research problem is to enable these models to successfully assist humans even when they make errors.  

\begin{figure}[t!]
\centering
\includegraphics[width=0.47\textwidth]{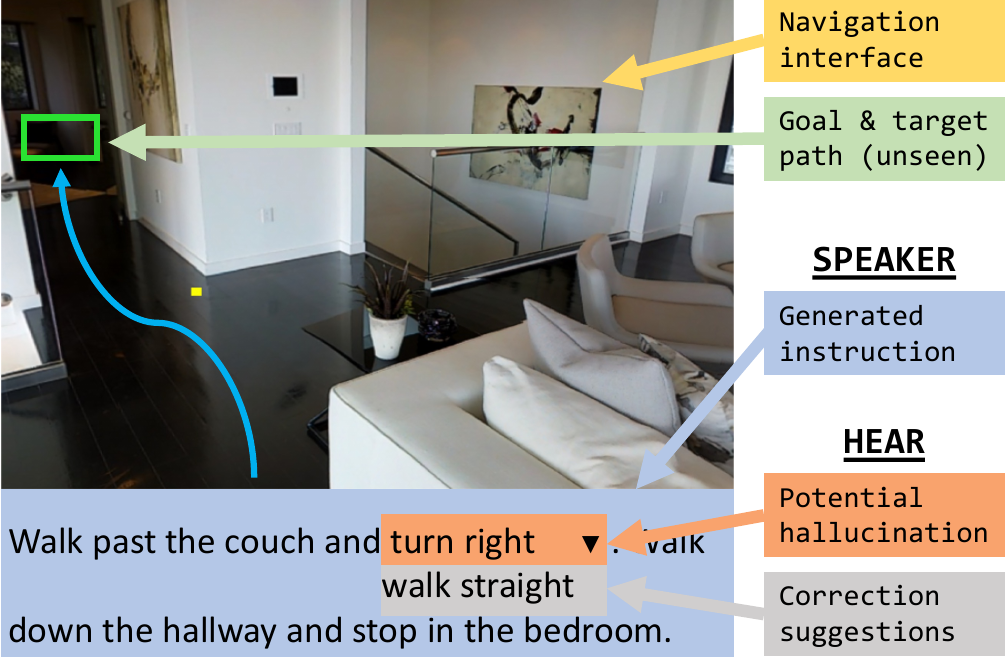}
\caption{\ourmodel detects errors in a navigation instruction and suggests corrections. It enables humans to avoid being misled and efficiently search the environment, leading to improved  performance. \looseness=-1}
\label{fig:problem}
\end{figure}

But how is it possible for a model to guide a human toward the right decisions when it cannot precisely specify what those decisions are?
This work demonstrates the feasibility of tackling this problem in a language-guided visual navigation setting. 
Concretely, we develop \ourmodel (\textbf{H}allucination D\textbf{E}tection \textbf{A}nd \textbf{R}emedy), a system that aids human navigation in 3D residential environments using potentially erroneous natural language instructions.
The key to the success of \ourmodel is its ability to communicate various types of uncertainty information to users. 
Specifically, \ourmodel can identify and highlight potential errors in an instruction, and suggest possible corrections.
This information prevents misdirection and narrows the search space for users, enabling them to navigate successfully even when given inaccurate instructions. 

To our best knowledge, our work presents the first study on the effects of uncertainty communication on human decision making in a long-horizon task. 
Although uncertainty communication has been identified as crucial for AI systems, very few studies have investigated how uncertainty information impacts human decisions. Previous studies have primarily focused on classification tasks rather than long-horizon tasks, and on numerical uncertainty (i.e., probability) rather than verbal uncertainties \citep{vodrahalli2022uncalibrated,nizridoes}. By demonstrating that presenting uncertainties leads to a substantial performance boost in this navigation task, we provide strong evidence to support the development of these features in sequential decision-making AI agents.

To build \ourmodel, we tackle the problem of detecting and classifying hallucinated phrases in visually grounded instructions. 
This problem is particularly challenging in the environments we study because of the realisticity and diversity of the visual scenes. 
Our solution involves training two vision-language models: one for hallucination detection and the other for classification (i.e., deciding whether a phrase should be deleted or replaced).  
We combine these models to identify hallucinations in an instruction, as well as score and rank potential corrections.
To train each model, we fine-tune a large vision-language model \cite{guhur2021airbert} with synthetically created data to optimize for a contrastive learning objective.
We introduce a practical methodology for generating synthetic data, combining rule-based approaches with large language models.\looseness=-1

We conduct an evaluation with 80 human users to measure the effectiveness of \ourmodel.
Our results demonstrate that incorporating HEAR improves user navigation outcomes.
Specifically, HEAR increases the likelihood of a user successfully reaching their destination by 13\% and reduces the average distance to the true destination by 29\%.
Analyzing human behavior reveals that by providing useful hints, \ourmodel motivates humans to put more effort into solving a task, leading to a higher success rate.\looseness=-1

Interestingly, our results suggest that the uncertainty communication capabilities of a system do not need to be flawless to boost user performance. 
The components of \ourmodel are all imperfect: the error detection and correction, and the instruction generation capabilities are all of reasonable quality, but not faultless. 
However, because these capabilities \textit{complement} one another, and complement the knowledge of the human user, they ultimately improve user decisions.

%% file: sections/related_work.tex
\paragraph{Grounded instruction generation.}
Grounded instruction generation involves creating language instructions for navigation in situated environments, evolving from simple settings \cite{anderson1991hcrc, goeddel2012dart, fried2017unified} to more complex, photo-realistic simulations \cite{fried2018speaker, kamath2022new, zhao2023cognitive}. 
Model-generated instructions can contain landmark errors (e.g., confusing a bathroom with a gym) and path errors (e.g., instructing a left turn instead of a right turn) \cite{wang2022less}. \citet{zhao2023cognitive} demonstrate a significant gap between the quality of model- and human-generated instructions.
However, their work is not concerned with error detection.

\paragraph{Uncertainty communication for human-AI collaboration.}
As AI-assisted decision-making has become the norm, it is imperative to investigate the influence of human cognitive biases on their perception of model-generated information \citep{rastogi2022deciding}. Several studies have questioned the necessity of probabilistic calibration, showing that presenting \textit{un}calibrated probabilities may improve human decisions c\cite{benz2023human, vodrahalli2022uncalibrated, nizridoes}. Other research proposes model designs to better calibrate human trust \cite{zhang2020effect, ma2023should, buccinca2021trust}. The experimental settings in all of these papers focus on classification tasks rather than long-horizon decision-making tasks, as explored in this work.

Regarding complementary performance in human-AI collaboration, \citet{bansal2021does} famously demonstrate that presenting model-generated explanations to humans does not enable human-AI teams to outperform individual entities. 
We present a contrasting result, showing that a complementary performance boost is possible with careful selection and presentation of model-generated information.

\paragraph{Hallucination detection.}
Neural text generation models produce hallucinations in textual domains \cite{kalai2023calibrated, muller2019domain, maynez2020faithfulness, durmus2020feqa, liu2021token} as well as multimodal domains \cite{wiseman2017challenges,rohrbach2018object,liu2024survey,chen2024multi}.
Hallucination detection has been explored, but primarily for machine translation \citep{dale2022detecting, xu2023understanding, wang2020exposure,zhou2020detecting} or summarization \citep{falke2019ranking, kryscinski2019evaluating,chen2021improving}.
Closest to our work is \citet{zhao2023hallucination}, who study this problem in a similar visual navigation setting. However, their model cannot provide correction suggestions, nor do they design user interfaces or perform evaluations with real human users.

%% file: sections/problem.tex
We consider the problem of generating language instructions to guide a human to follow an intended route in an environment. 
The concrete goal is to build a \textit{speaker model} $S(\instr \mid \traj)$, which takes an intended route $\traj$ as input and generates a corresponding language instruction $\instr$ as output (\autoref{fig:problem}).
The instruction $\instr = (w_1, \ldots, w_n)$ is a sequence of words (e.g., {\fontfamily{cmss}\selectfont ``Walk past the couch and turn right. Walk down the hallway and stop in the bedroom.''}).
The route $\traj = (\vec o_1, a_1, \ldots, \vec o_l, a_l)$ is a sequence of observations and actions, where each observation is a collection of RGB images that capture the view at a location, and each action represents a transition from one location to another.
The speaker is evaluated through an \textit{instruction-following} task, in which a human user receives an instruction generated by the speaker and follows it in the corresponding environment.
Success is achieved if the user reaches the final location along the intended route.

To simulate this problem, we employ the Matterport3D simulator and Room-to-Room (R2R) dataset \cite{anderson2018vision} for model training and human experiments. 
Matterport3D is a photorealistic simulator that features images taken from various real residential buildings.
The R2R dataset contains pairs of route and language instruction.
The instructions contain more than 7,000 object and direction phrases.

We follow \citet{zhao2023cognitive} to train a T5-based \citep{raffel2020exploring} speaker model. 
The instructions generated by this model often contain object or directional phrases that are inconsistent with the scenes along the intended route.
We refer to such phrases as \textit{hallucinations}.
We categorize hallucinations into two types: \textit{intrinsic hallucination} is a phrase that needs to be replaced because it inaccurately describes an observation or action (e.g., 
{\fontfamily{cmss}\selectfont an instruction says ``Walk past the reception desk and out the door on the \textcolor{purple}{right}''}, but on the intended route, the door is on the left);
\textit{extrinsic hallucination} is a phrase that needs to be removed because it does not have a correspondence on the input route (e.g., {\fontfamily{cmss}\selectfont ``Walk through the office and out of the office. \textcolor{purple}{Walk into the hallway and turn left}''}, where the second sentence describes a path that does not exist in the environment).
Upon inspecting 40 sample instructions generated by our speaker, we find that 67.5\% of them have hallucinations, and that 20.9\% of all the object and direction phrases are hallucinations.
\looseness=-1

%% file: sections/model.tex
\begin{figure*}[t!]
\centering
\includegraphics[width=\textwidth]{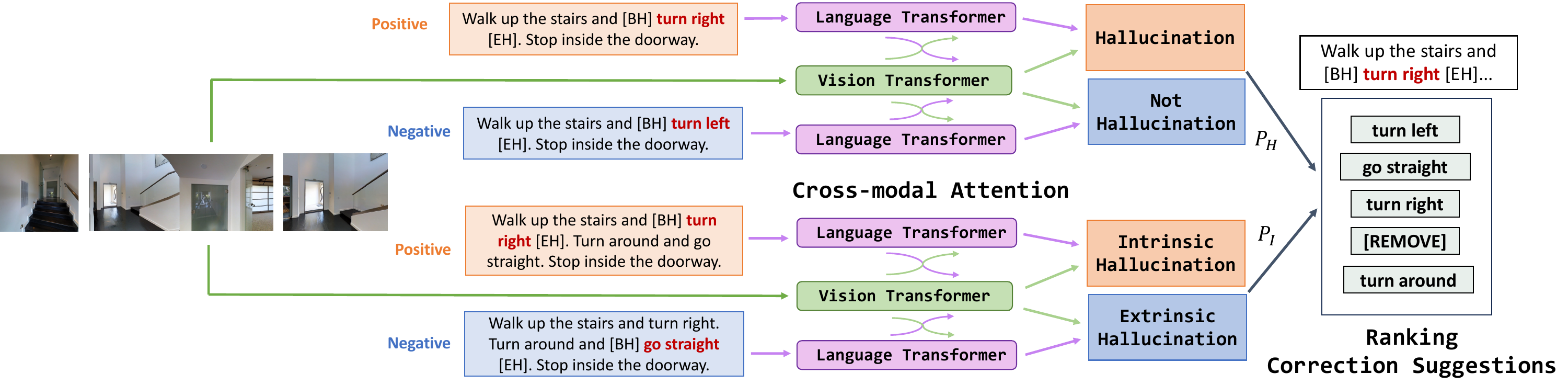}
\caption{Our hallucination detection model (top) and hallucination type classification model (bottom). Each model takes a language instruction and a visual route as input and predicts a binary label. 
For hallucination detection, the label is whether a phrase is a hallucination. For hallucination-type classification, the label is whether a hallucination is extrinsic (needed to be replaced) or extrinsic (needed to be removed). Each model is built on top of a pre-trained vision-language model and is fine-tuned using contrastive learning. The first model is used to decide which phrases to highlight in an instruction, and the two models are combined to score and rank possible corrections. 
\looseness=-1}
\label{fig:model}
\end{figure*}

In this section, we introduce \ourmodel, which augments a speaker model by enabling it to (i) highlight potential hallucinations in an instruction and (ii) produce a list of plausible corrections for each hallucination. 
We expect that (i) would help a user avoid being misled into incorrect regions, while (ii) would reduce the effort required to locate the correct region.
We build two models (\autoref{sec:hallucination_detection_model}, \autoref{sec:correction_suggestion_model}, illustrated in \autoref{fig:model}) to generate these pieces of information and design an interface to effectively convey them to users (\autoref{sec:communication_interface}). \looseness=-1

\subsection{Hallucination Detection}
\label{sec:hallucination_detection_model}

The hallucination detection model predicts hallucinations in an instruction.
We adopt the model from \citet{zhao2023hallucination} but train it on a different training set so that it can detect  \textit{phrases} instead of just tokens as in the original work. 

We frame the hallucination detection problem as a binary classification task: given an input $\vec x = (\traj, \instr, i, j)$ consisting of a route $\traj$, an instruction $\instr$, and token indices $i, j \in \{1, \cdots, n\} (i \leq j)$, decide whether the phrase $\instr_{i:j} = (w_i, w_{i+1}, ..., w_j)$ is a hallucination (more specifically, whether it should be replaced or removed to make $\instr$ consistent with $\traj$). 
For example, in the instruction shown in \autoref{fig:problem}$, \instr_{6:7}$ is predicted to be a hallucination. 
We use a combination of a POS tagger\footnote{\url{https://spacy.io}} and GPT-3.5-turbo to identify the phrases to be classified.

Our model is a classifier $P_H (y=1 | \vec x = (\traj, \instr, i, j))$  that is fine-tuned from the Airbert model \cite{guhur2021airbert}---a vision-language model pre-trained on a large corpus of captioned household scenes collected from AirBnB. 

For each instruction, we wrap the phrases to be classified between a pair of special tokens (\texttt{[BH]} and \texttt{[EH]}).
For example, if $\vec w_{i:j}$ is classified, the instruction becomes
$\left[~ w_1, \dots, \texttt{[BH]}, w_i, ..., w_{j}, \texttt{[EH]}, \dots, w_{n} ~\right]$. The model takes as input this annotated instruction and the visual route and outputs a score $s(\vec x)$.
The hallucination confidence is calculated as $P_H (\vec x) = \sigmoid(s(\vec x))$, where $\sigmoid$ is the sigmoid function. 
The model is trained with a contrastive objective \citep{majumdar2020improving} on pairs of positive and negative examples (described in \autoref{sec:dataset_creation}). 

\subsection{Correction Suggestion}
\label{sec:correction_suggestion_model}

For each phrase $\instr_{i:j}$ classified as hallucination by $P_H$, we compute the top-$K$ correction suggestions.
To do so, we first generate a set of candidate corrections $\{\correctphrase^m_{i:j}\}_{m=1}^M$ (this procedure will be described in \autoref{sec:dataset_creation}). 
For example, in Figure \ref{fig:problem}, $\{\correctphrase^m_{6:7}\}$ is \{{\fontfamily{cmss}\selectfont turn right}, {\fontfamily{cmss}\selectfont walk straight}\}. 
A special token {\fontfamily{cmss}\selectfont [REMOVE]} represents the deletion of the phrase.  
We train a hallucination-type classification model, which allows us to rank these candidates and choose the top $K$.
\looseness=-1

\paragraph{Ranking suggestions.}
As mentioned in \autoref{problem}, we categorize hallucinations into two types: \textit{intrinsic} and \textit{extrinsic}.
Let $z_{\vec x}$ denote the hallucination type of a phrase $\vec x$; $z_{\vec x} = 1$ if $\vec x$ is an intrinsic hallucination.
We learn a binary classifier to estimate $P_I(z = 1 \mid \vec x, y_{\vec x} = 1)$ where $y_{\vec x} = 1$ indicates that $\vec x$ is a hallucination.
Let $\vec x = (\vec r, \vec w, i, j)$ and $\hat{\vec x}$ be the corrected version of $\vec x$ obtained by replacing $\vec w_{i:j}$ with a candidate correction $\hat{\vec w}_{i:j}$.
We compute a score $R(\hat{\vec x})$ for every candidate (the higher is the better).  
We consider two cases.
If $\hat{\vec x}$ indicates a replacement, we define $R(\hat{\vec x})$ as:
\begin{align}
     P_I(z = 1 \mid \vec x, y_{\vec x} = 1) \cdot P_H(y = 1 \mid \hat{\vec x})
\end{align} where the first term computes how likely $\vec x$ necessitates a replacement, while the second term captures how good the proposed replacement $\hat{\vec x}$ is. 
If $\hat{\vec x}$ indicates a deletion, 
we set $R(\hat{\vec x}) = P_I(z = 0 \mid \vec x, y_{\vec x} = 1)$, which estimates the probability that $\vec x$ is an extrinsic hallucination (thus requiring deletion).
\looseness=-1

\paragraph{Hallucination type classification.}
The model $P_I$ uses the same model architecture and is trained in a similar fashion as the hallucination model $P_H$.
However, it solves a different problem: determining the type of a hallucination rather than identifying whether a phrase is a hallucination.
This is achieved by training on a different dataset, as described in \autoref{sec:dataset_creation}.

\subsection{Dataset Creation}
\label{sec:dataset_creation}

To train the models described in previous sections, we construct training datasets with positive and negative examples, defined by the specific classification problem. 
We also create a set of candidate corrections for each predicted hallucination. 
As human-labeled training data is costly to obtain, we synthetically create training data by taking human-generated instructions in the R2R training set and perturbing them using rule-based procedures and GPT models.

\paragraph{Training data for hallucination detection.}
For this problem, the negative examples are instructions from the R2R training set \citep{anderson2018vision}, which are assumed to contain no hallucinations. 
To create a positive example from a negative example denoted by $\vec x^- = (\traj, \instr^-, i, j)\}$, we perturb the instruction $\vec w^-$ in various ways.
Following \citet{zhao2023hallucination}, we focus on three types of intrinsic hallucinations: room, object, and direction. 
We create a room hallucinations by replacing a room phrase with another randomly chosen from a pre-composed list, and generate an object hallucination by replacing an object phrase with another that appears in the same instruction.
For directions, since one can be expressed in various ways (e.g. {\fontfamily{cmss}\selectfont go straight} is the same as {\fontfamily{cmss}\selectfont proceed forward}), we leverage GPT-3.5-turbo to modify them, using the following prompt (the few-shot examples are not shown for brevity; the full prompt is in \autoref{app:dataset_creation}):

\noindent\fbox{%
    \parbox{\columnwidth}{%
\textbf{SYSTEM:}
Find a directional word/phrase in the original instruction, and substitute it with a completely different directional word/phrase, so a person following the modified instruction would go in a different direction from the original instruction. Craft three modified instructions for each original instruction, and utilize the <s></s> tag pair to highlight the directional word/phrase you've modified in both the original and modified instructions.

\textbf{Input:} {\fontfamily{cmss}\selectfont Walk out of the bedroom and turn left.} 

\textbf{Output:}
{\fontfamily{cmss}\selectfont <original1> walk <s> \colorbox{blue!30} {out of} </s> the bedroom and turn left . </original1> 

 <modified1> walk <s> \colorbox{orange!50}{around} </s> the bedroom and turn left . </modified1>}

    }%
}

Meanwhile, an extrinsic hallucination in an instruction is constructed by inserting a sentence taken from the same or a different instruction into a randomly selected beginning-of-sentence location within the instruction.

Multiple hallucinations are created within an instruction, but only one is wrapped by the \texttt{[BH]} \texttt{[EH]} tags for classification.
We also add hallucinations to the negative example, but ensure that the span enclosed by \texttt{[BH]} \texttt{[EH]} is not a hallucination.

\paragraph{Training data for hallucination-type classification.}

For this dataset, both the positive and negative examples contain hallucinations, but the enclosed spans in the positive examples are intrinsic hallucinations, while those in the negative examples are extrinsic hallucinations.
We apply the approach used in the detection problem to synthesize hallucinations.

\paragraph{Generating sets of candidate corrections.}
We generate a set of candidate corrections for each predicted hallucination. 
The candidate corrections for a room or an object hallucination are all the rooms and objects provided by the Matterport3D simulator. 
For directions, we ask GPT-4 to generate candidates, using the following prompt (the few-shot examples are not shown; the full prompt is in \autoref{app:dataset_creation}):
\noindent\fbox{%
    \parbox{\columnwidth}{%
\textbf{SYSTEM:}
Find directional words/phrases in the instruction and use <original> </original> tags to mark them, and list all the possible substitutions to change the meaning completely with <modified> </modified> tags, so that a person following the substituted instruction would go in a different direction from the original instruction. Use <sep> to separate each substitution, and do not mark the nouns.

\textbf{Input:} {\fontfamily{cmss}\selectfont Walk out of the bedroom and turn left. }

\textbf{Output:} {\fontfamily{cmss}\selectfont walk <original1> \colorbox{blue!30}{out of} </original1> <modified1> \colorbox{orange!50}{into} <sep> \colorbox{orange!50}{around} <sep> \colorbox{orange!50}{to the left of} <sep> \colorbox{orange!50}{to the right of} </modified1> the bedroom and <original2> \colorbox{blue!30}{turn left} </original2> <modified2> \colorbox{orange!50}{go straight} <sep> \colorbox{orange!50}{turn right} <sep> \colorbox{orange!50}{turn around} </modified2> . }
    }%
}

On average, we generate 47.6 candidates for each room or object hallucination and 5.9 candidates for each direction hallucination.

\subsection{Designing Communication Interface}
\label{sec:communication_interface}

We build on top of the interface developed by \citet{pangea2021} and \citet{zhao2023cognitive} which allows a human to follow a language instruction to interact with a Matterport3D environment.
We augment the interface to display highlights and suggestions for potential hallucinations. 
This section discusses our design principle; more details and a visualization of the interface are given in \autoref{app:human_evaluation}. 

Our system generates a lot of information that can potentially be communicated to users.
Deciding what piece of information to present and how to present it is vital to the success of the system.
We choose not to present model probabilities to users because they can be miscalibrated and even if they are, different people might interpret them differently \citep{vodrahalli2022uncalibrated}. 
Instead, we convey binary predictions of hallucinations through highlights. 
To do so, we select a decision threshold for the hallucination detection model to maximize its F-1 score on a manually annotated development set.
If all phrases in a clause are highlighted, we simply highlight the entire clause and treat the clause as a single hallucination.
For each instruction, we highlight at most three hallucinations predicted by the model, which is approximately the average number of hallucinations in an instruction detected by our human annotators.

For suggestions, because their presence can be overwhelming, we display them only when the user deliberately seeks them out.
Initially, the user sees only the instruction (potentially with hallucination highlights).
We instruct them to click on a highlighted phrase if they also suspect it to be a hallucination and want to view possible corrections.  
If that happens, a drop-down menu will appear, displaying the top three suggestions in descending order by the score produced by our ranking models.
The user can click on a suggestion to apply it to the instruction, which closes the drop-down menu.
We explicitly instruct users to correct the instruction to encourage them to consider the suggestions.

A complication we encounter is to decide how much information about the true final location should be revealed to the users.
If users do not know the true final locations, they cannot correct the instructions. 
However, if the location is completely revealed to them, the influence of the instructions on their behavior is significantly weakened, undermining the purpose of our study.
To address this issue, we introduce a \textit{Check} button, which enables the human to verify whether they have reached the final location.
The button enables users to correct instructions while also retaining their reliance on instructions.
In addition, analyzing user button usage uncovers interesting insights about their behavior.

%% file: sections/experiments.tex
The questions that we aim to answer are:
\begin{enumerate}[nolistsep]
    \item[(Q1)] Can \ourmodel reliably detect hallucinations and provide reasonable suggestions?
    \item[(Q2)] Does providing hallucination highlights and suggesting corrections improve human navigation performance?
    \item[(Q3)] What are the effects of highlights and suggestions on human behavior?
\end{enumerate}

To answer Q1, we evaluate \ourmodel intrinsically with human-annotated data.
To answer Q2 and Q3, we conduct a human evaluation with various systems, including ablated versions of \ourmodel and an oracle human-based system.

\paragraph{Data.} 
To train the hallucination detection model, we synthetically generate a training set with 164,939 pairs of positive and negative examples (\autoref{sec:dataset_creation}), which are created from the Room-to-Room (R2R) \cite{anderson2018vision} train set (4,675 routes, each route has 3 human-annotated instructions). 
To train the hallucination type classification model, we generate 117,357 pairs of positive and negative examples, created from the R2R train set.

For both evaluations, we first use a speaker model (\autoref{problem}) to generate instructions describing routes from the R2R validation seen split.
For intrinsic evaluation and model selection, we randomly select and annotate 40 routes from the split as our \textit{Dev Set}.
For human evaluation, we use the 75 test routes from previous work \cite{zhao2023cognitive, zhao2023hallucination} as our \textit{Test Set}. 
There is no overlap between the Dev Set and the Test Set.

\subsection{Intrinsic Evaluation: Hallucination Detection and Correction Suggestion}
\label{sec:instrinsic-eval}

\paragraph{Annotation.} 
We manually annotate hallucinations in the instructions generated by the speaker model, with mutual agreement from two of the authors. 
We also annotate corrections for those spans that we label hallucinations.
In the end, we create intrinsic evaluation datasets consisting of 376 examples from the \textit{Dev Set} for model selection; and 625 examples from the \textit{Test Set} for testing, as well as used by the Oracle system for human evaluation (\autoref{sec:extrinsic_eval}). \looseness=-1

\paragraph{Systems.}
We implemented the following approaches (detailed hyperparameters in \autoref{app:models}): 
\begin{enumerate}[label=(\alph*),nolistsep]
    \item \textit{\ourmodel} is our final system described in \autoref{sec:hallucination_detection_model}, \autoref{sec:correction_suggestion_model}, and \autoref{sec:dataset_creation}.
 \item \textit{HEAR-SameEnvSwap} is similar to \ourmodel but the strategy to create room and object hallucinations is slightly different. Instead of following the procedure described in \autoref{sec:dataset_creation}, we swap objects and rooms with those in the same environment (more details in \autoref{app:model_variants}).
 \item \textit{One-stage \ourmodel} combines hallucination detection and type classification into a single model (more details in \autoref{app:model_variants}). This model can directly score each correction suggestion. 
 \item \textit{Random} samples a label uniformly at random among all possible labels, where the labels are \{yes, no\} for hallucination detection, and are the set of all possible 3-element subsets of the candidate set for correction suggestion.

\end{enumerate}

\paragraph{Metrics.} We compute \textit{macro-averaged F-1} for hallucination detection and compute \textit{Recall@3} for correction suggestion, which is the empirical chance that the gold correction appears in the top-3 suggestions ranked by a system. 
\looseness=-1

\paragraph{Main results (\autoref{table:intrinsic_eval}).}

\begin{table}[t]
\centering
\small
\begin{tabular}{@{}lcccc@{}}
\toprule
            & \multicolumn{2}{c}{Dev} & \multicolumn{2}{c}{Test} \\ \cmidrule(lr){2-3} \cmidrule(lr){4-5}
System            & F-1      & R@3     & F-1       & R@3     \\ \midrule
Random   &   42.6   &    47.8      &   43.8   &    50.4      \\ 
\ourmodel-SameEnvSwap & \textbf{64.8}     & 75.0         & \textbf{69.1}      & 78.7         \\
One-stage \ourmodel   & 62.8     & 82.7         & 60.9      & \textbf{86.2}         \\
\ourmodel (final)       & 63.4     & \textbf{88.4}         & 66.5      & 70.6         \\

\bottomrule
\end{tabular}
\caption{Intrinsic evaluation of \ourmodel and our baseline systems. The decision threshold for each system is selected to maximize the F-1 score on the Dev Set. R@3 computes how often the top-3 correction suggestions contain the gold annotated correction. 
\looseness=-1}
\label{table:intrinsic_eval}
\end{table}

All the learned models substantially outperform the random baseline. 
In particular, the R@3 metrics of these models are in the range of 70-90\%, showing that they have a high potential to aid humans.

The results in hallucination detection show a clear trend, HEAR-SameEnvSwap is the best model in terms of F-1 score, followed by \ourmodel and finally one-stage \ourmodel.
This indicates that the data-creation strategy in the HEAR-SameEnvSwap training set is beneficial. 
Meanwhile, the performance of one stage \ourmodel is low, possibly because it has twice as few parameters as the other two models.
The results in correction suggestion recall are more nuanced: \ourmodel is best on Dev but one-stage \ourmodel is superior on Test.
HEAR-SameEnvSwap outperforms others in hallucination detection, but its underperformance in correction suggestion indicates that the probabilities output by its hallucination detection module are not reliable.

Considering the average of F-1 and R@3, \ourmodel is the best performing model on the Dev set. Therefore, we select it for evaluation with human users.
\looseness=-1

\begin{figure*}[ht]
\centering
\includegraphics[width=\textwidth]{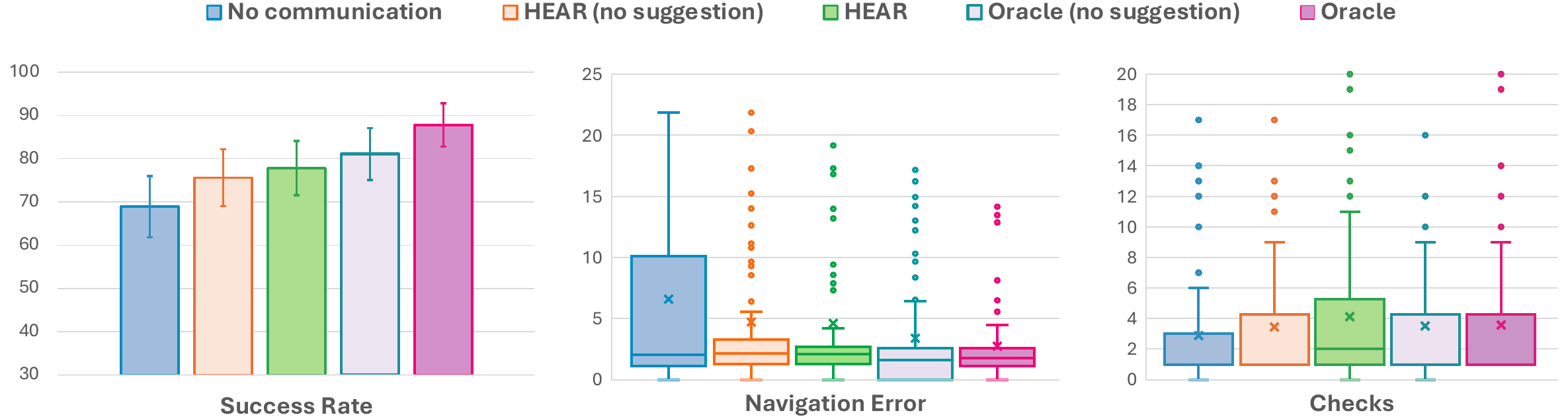}
\caption{Performance measured by success rate (SR ↑) and navigation error (DIST ↓), and the number of check-button clicks recorded when human users perform navigation tasks with different assistant systems. 
\ourmodel improves user navigation performance and is competitive with the two Oracle systems.
The error bars for SR represent 85\% confidence intervals.
For DIST and Checks, the ``x'' marks the mean, the line inside a bar marks the median, and the box represents the interquartile. \autoref{tab:main_result} shows the corresponding results in table format.  \looseness=-1} 
\label{fig:human_sr}
\end{figure*}

\subsection{Extrinsic Evaluation with Human Followers}
\label{sec:extrinsic_eval}

\paragraph{Setup.}
We evaluate five systems:
\begin{enumerate}[label=(\alph*),nolistsep]
    \item \textit{No communication} only tells the user that the instruction may be imperfect. It does not provide highlights and suggestions, and is similar to the system in \citet{zhao2023cognitive}.
    \item \textit{\ourmodel (no suggestion)} tells the user that the instructions can be imperfect, highlights potential hallucinations, and tells the user that those phrases are potential errors. It does not provide suggestions. This system is similar to \citet{zhao2023hallucination}.
    \item \textit{\ourmodel} is our final system, which adds to (b) the ability to suggest the top three corrections for each predicted highlight. We choose to present the top three suggestions to balance the system's recall performance with user mental load.\looseness=-1
    \item \textit{Oracle (no suggestion)} is similar to (b) but highlights are annotated by the authors.
    \item \textit{Oracle} is similar to (c), but highlights and corrections are annotated by the authors. It displays two instead of three candidate suggestions: the original phrase and the gold correction.
\end{enumerate}

We evaluate each system on 18 routes randomly chosen from the Test Set.
For each route and each system, we recruit five human users using Amazon Mechanical Turk and ask them to follow the instruction generated by the system to describe the route.
Users are paid \$4.10 for each session, which involves performing 7 navigation tasks and takes on average 19 minutes to complete.
One of the tasks is a quality-control task that appears in every session.
We analyze only sessions in which the user passes this task.
After completing a session, users can provide feedback on the system.
We ensure that each user encounters each route only once to prevent them from memorizing it.
In total, we recruit 80 users and evaluate 525 navigation tasks. 
\looseness=-1

\paragraph{Metrics.} We evaluate navigation performance using standard metrics of the R2R task: 
\begin{enumerate}[label=(\alph*),ref=\alph*,nolistsep]
    \item Success rate (\textit{SR $\uparrow$}): fraction of examples in which the user's final location is within 3m of the true goal;
    \item Navigation error (\textit{DIST $\downarrow$}): distance between the user's final location and the true goal.
\end{enumerate}
After a user has finished navigating, we ask for their subjective judgements about the route and the instruction, specifically: 
\begin{enumerate}[label=(\alph*),ref=\alph*,nolistsep]
\item Is the instruction easy to follow?
\item Are you confident the path you followed is the intended path?
\item Is the task mentally demanding?
\end{enumerate} For each question, we use 5-point Likert scale to ask for a rating on the affirmative statement (e.g., {\fontfamily{cmss}\selectfont I am confident that I traversed the path that the AI system tried to describe}). 

\begin{table}[t]
\setlength{\tabcolsep}{2pt}
\small
\centering
\begin{tabular}{@{}lccc@{}}
\toprule
& Easy to& Confident& Mental \\
System & follow? $\uparrow$ & on actions? $\uparrow$ & burden? $\downarrow$ \\ \midrule
No communication            &     3.7     &   3.8    &    3.6     \\
\ourmodel (no suggestion)             &     3.5     &   3.9    &   \textbf{3.5}      \\
\ourmodel   &   4.0      &   ~~~\textbf{4.2} $^{\ddagger}$    &     \textbf{3.5}    \\
Oracle (no suggestion)  &     3.9     &    3.8   &    3.6     \\
Oracle  &     ~~\textbf{4.1}$^{\dagger}$     &  ~~4.1$^{\dagger}$  &    3.7     \\ \bottomrule
\end{tabular}
\caption{User subjective ratings of systems after completing navigation sessions. 
The symbols $^{\ddagger}$ and $^{\dagger}$ indicate results that are significantly higher than those of the ``No communication'' system in the first row, with $p < 0.004$ (Bonferroni correction for 12 tests comparing 4 systems with ``No communication'') and $p < 0.05$, respectively, as determined by a two-related-sample t-test.
\looseness=-1}
\label{table:human_subjective}
\end{table}

\begin{figure*}[t]
\centering
\begin{subfigure}{0.45\textwidth}
\includegraphics[width=0.8\textwidth]
{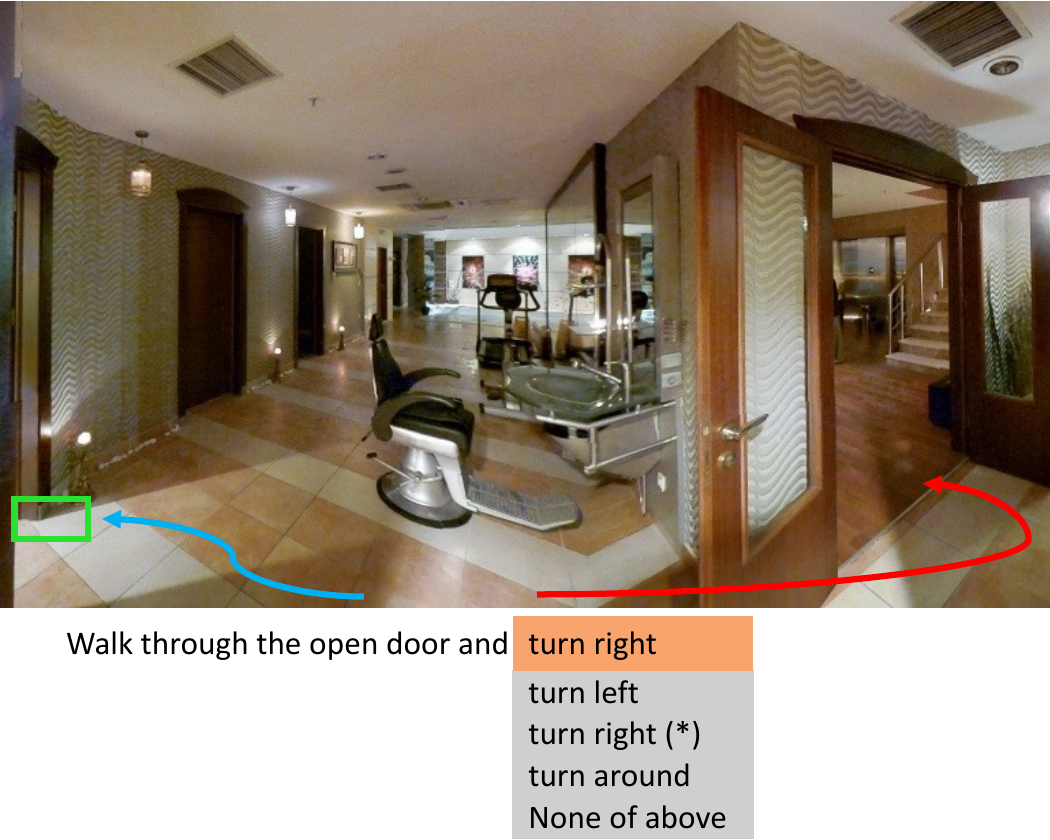}
\centering
\caption{Highlights and suggestions directs a user to correctly make a left turn (blue). With only highlights, another user mistakenly turns right (red).\looseness=-1}
\label{fig:qual_alternative_better_highlight}
\end{subfigure}
\hfill
\begin{subfigure}{0.45\textwidth}
\includegraphics[width=0.8\textwidth]{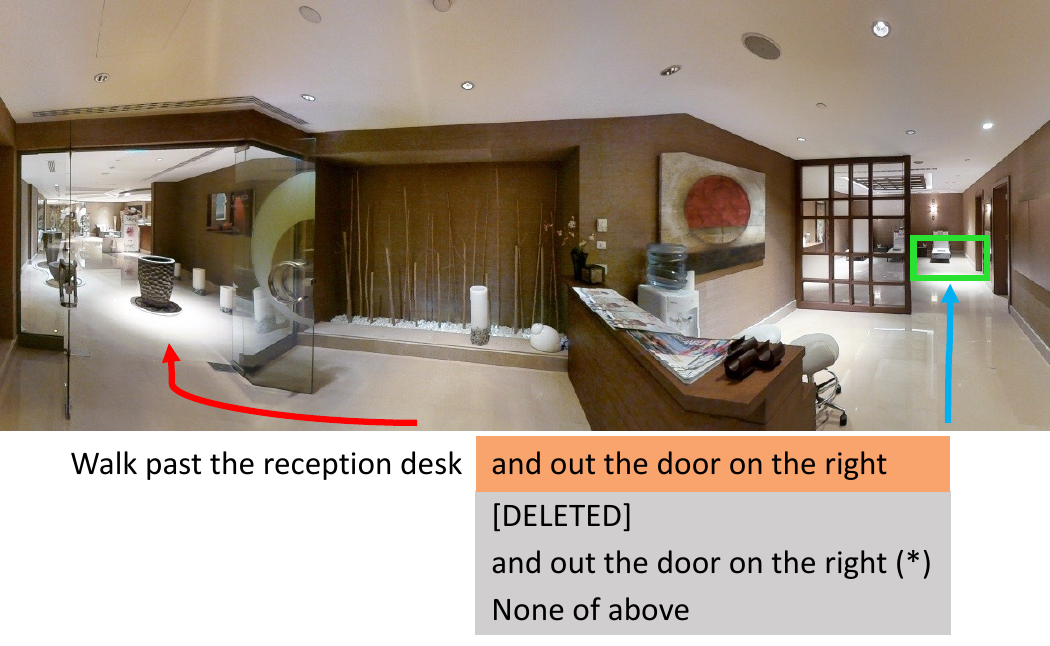}
\centering
\caption{
A user successfully reaches the destination solely with the highlight (blue), while another fails upon receiving additional suggestions (red). While the highlight and the top suggestion ([delete]) are incorrect, they appear to reinforce each other, making the user believe that the highlight is correct and go in the alternative direction.
}
\label{fig:qual_highlight_better_alter}
\end{subfigure}
\caption{Example success and failure cases of \ourmodel (more in \autoref{app:qualitative_examples}). \looseness=-1}
\label{fig:qual_examples}
\end{figure*}

\paragraph{\ourmodel enhances user navigation performance.}
As seen in \autoref{fig:human_sr}, compared to no communication, simply highlighting potential errors using \ourmodel increases user success rate (+6.7\%) and decreases navigation error (-1.9m).
These results confirm that error highlights can effectively compensate for the deficiencies of the instruction generation model.
A user described the effects of highlights as follows: ``\textit{highlights help me know if the instructions were going to be wrong. It made it easier to know where to go back to and retrace steps in order to go to the right place}''.
User performance is further improved with suggestions generated by \ourmodel (+2. 2\% in SR and -0.1m in DIST). 
\autoref{fig:qual_alternative_better_highlight} shows an example where a user who is provided with both highlights and suggestions successfully reaches the target destination, whereas another user who is shown only highlights does not.
\looseness=-1

Another notable pattern, shown in \autoref{fig:human_sr} (middle), is that adding highlights and suggestions substantially decreases the variance of the navigation error.
This indicates that highlights and suggestions effectively reduce the search space of the users.

\paragraph{\ourmodel receives favorable subjective ratings.} As shown in \autoref{table:human_subjective}, users find the instructions generated by \ourmodel (and Oracle systems) easier to follow and report greater confidence in their actions. 
Despite being asked to correct errors in the instructions, users do not report a significant increase in mental load.

\paragraph{\ourmodel improves user persistence in completing tasks.}
\autoref{fig:human_sr} (rightmost) shows that users, on average, use the Check button more often when provided with highlights and suggestions. 
This result suggests that these features incentivize users to make more attempts to solve the task and consequently become more successful. 
We hypothesize that by suggesting possibilities for exploration, users can avoid blind searches, making them more willing to invest effort.
In contrast, without highlights and suggestions, users lack direction and may give up more quickly. 
They may perceive an entire instruction as incorrect and believe that the correct instruction could be entirely different from the current one, leading them to feel there is no hope in searching without further clues. 
\looseness=-1

\paragraph{Better highlights and suggestions further improve user performance.}
\autoref{fig:human_sr} shows that users benefit from a better hallucination detection model; they achieve a higher success rate (+5.5\%) and a smaller navigation error (-1.3 m) when Oracle highlights are given, compared to when \ourmodel highlights are presented.

User performance is also enhanced when using an improved correction suggestion model: +10.0\% in success rate and -1.9m in navigation error when using Oracle suggestions compared to when using \ourmodel suggestions. \autoref{fig:qual_highlight_better_alter} illustrates how a user is misled by incorrect highlights and suggestions.

%% file: sections/conclusion.tex
We present a novel approach to enhance human task performance by effectively communicating model uncertainties. By encouraging users to refine AI-generated solutions, our approach offers an alternative to the conventional method that focuses on directly improving AI autonomous capabilities while overlooking human capabilities.
To fully unlock the potentials of AI technologies, we advocate for viewing AI systems not as independent problem solvers, but as assistants and collaborators of humans.
\looseness=-1

While our research primarily addresses language-guided visual navigation, the insights gained are broadly applicable to other vision-language tasks. Specifically, we have demonstrated that: (i) it is feasible to generate meaningful error highlights and correction suggestions for vision-language models, and (ii) presenting these highlights and suggestions to human users can improve their decision-making. Moreover, our methods for creating synthetic errors and correction suggestions using rules and large language models are generalizable to various contexts.

%% file: sections/limitations.tex
Due to cost constraints, the scale of our human evaluation is limited.
We prioritize having more annotators evaluate each route over having more routes. 
Furthermore, the assessment of cognitive load in the human evaluation study is not sufficiently robust; we plan to administer other schemes, such as the NASA Task Load Index \citep{hart2006nasa}, in future work.
\looseness=-1

Before using the navigation interface, users watch a video tutorial that explains the components of the interface and the associated questions. However, this could be improved by incorporating a warm-up practice session to help users become more familiar with the interface.

Another limitation of our human study is that we cannot determine how much of the performance improvement can be attributed to specific highlights and their associated correction suggestions, as task performance is assessed solely based on how close users are to the true final location.
Additionally, we do not record the time when the Check button is pressed, which prevents us from analyzing the distribution of button presses throughout a navigation process.
\looseness=-1

%% file: sections/appendix.tex
\subsection{GPT for Dataset Creation}
\label{app:dataset_creation}

The following prompt is given to GPT-3.5-turbo to create direction hallucinations in instructions (\autoref{sec:dataset_creation}):

\noindent\fbox{%
    \parbox{\columnwidth}{%
    \small
\textbf{Input:} Walk out of the bedroom and turn left. Walk into the kitchen and stop by the counter.

\textbf{Output:}
(1) <original1> walk out of the bedroom and <s>turn left</s> . walk into the kitchen and stop by the counter . </original1> 
<modified1> walk out of the bedroom and <s>turn right</s> . walk into the kitchen and stop by the counter . </modified1>

(2) <original2> walk <s>out of</s> the bedroom and turn left . walk into the kitchen and stop by the counter . </original2> 
<modified2> walk <s>around</s> the bedroom and turn left . walk into the kitchen and stop by the counter . </modified2>

(3) <original3> walk out of the bedroom and turn left . walk <s>into</s> the kitchen and stop by the counter . </original3> 
<modified3> walk out of the bedroom and turn left . walk <s>out of</s> the kitchen and stop by the counter . </modified3>

\textbf{Input:} Walk straight and turn left. Walk down the hallway and stop in the first doorway on your left.

\textbf{Output:}
(1) <original1> walk straight and turn left . walk <s>down</s> the hallway and stop in the first doorway on your left . </original1> 
<modified1> walk straight and turn left . walk <s>up</s> the hallway and stop in the first doorway on your left . </modified1>

(2) <original2> walk straight and turn left . walk down the hallway and stop in the first doorway <s>on your left</s> . </original2> 
<modified2> walk straight and turn left . walk down the hallway and stop in the first doorway <s>to your right</s> . </modified2>

(3) <original3> walk straight and turn right . walk down the hallway and stop in the <s>first</s> doorway on your left . </original3> 
<modified3> walk straight and turn right . walk down the hallway and stop in the <s>second</s> doorway on your left . </modified3>

\textbf{Input:} Exit the bathroom. Walk forward and go down the stairs. Stop four steps from the bottom.

\textbf{Output:}
(1) <original1> exit the bathroom . walk <s>forward</s> and go down the stairs . stop four steps from the bottom . </original1> 
<modified1> exit the bathroom . walk <s>backward</s> and go down the stairs . stop four steps from the bottom . </modified1>

(2) <original2> <s>exit</s> the bathroom . walk forward and go down the stairs . stop four steps from the bottom . </original2> 
<modified2> <s>enter</s> the bathroom . walk forward and go down the stairs . stop four steps from the bottom . </modified2>

(3) <original3> exit the bathroom . walk forward and go down the stairs . stop four steps from the <s>bottom</s> . </original3> 
<modified3> exit the bathroom . walk forward and go down the stairs . stop four steps from the <s>top</s> . </modified3>

\textbf{Input:} walk through open door, turn left, walk toward fireplace turn right, stop outside doorway.

\textbf{Output:}
(1) <original1> walk through open door , turn left , walk toward fireplace turn right , stop <s>outside</s> doorway . </original1> 
<modified1> walk through open door , turn left , walk toward fireplace turn right , stop <s>inside</s> doorway . </modified1>

(2) <original2> walk through open door , <s>turn left</s> , walk toward fireplace turn right , stop outside doorway . </original2> 
<modified2> walk through open door , <s>go straight</s> , walk toward fireplace turn right , stop outside doorway . </modified2>

(3) <original3> walk through open door , turn left , walk <s>toward</s> fireplace turn right , stop outside doorway . </original3> 
<modified3> walk through open door , turn left , walk <s>away from</s> fireplace turn right , stop outside doorway . </modified3>

    }%
}

The following prompt is given to GPT-4 to generate candidate direction corrections (\autoref{sec:dataset_creation}):

\noindent\fbox{%
    \parbox{\columnwidth}{%
    \small

\textbf{SYSTEM:}
Find directional words/phrases in the instruction and use <original> </original> tags to mark them, and list all the possible substitutions to change the meaning completely with <modified> </modified> tags, so that a person following the substituted instruction would go in a different direction from the original instruction. Use <sep> to separate each substitution, and do not mark the nouns.

\textbf{Input:} Walk out of the bedroom and turn left. Walk into the kitchen and stop by the counter.

\textbf{Output:}
walk <original1> out of </original1> <modified1> into <sep> around <sep> to the left of <sep> to the right of </modified1> the bedroom and <original2> turn left </original2> <modified2> go straight <sep> turn right <sep> turn around </modified2> . walk <original3> into <original3> <modified3> out of <sep> pass </modified3> the kitchen and <original4> stop by <original4> <modified4> walk pass <sep> walk away from </modified4> the counter .

\textbf{Input:} Walk straight and turn left. Walk down the hallway and stop in the first doorway on your left.

\textbf{Output:}
<original1> walk straight </original1> <modified1> turn left <sep> turn right <sep> turn around </modified1> and <original2> turn left </original2> <modified2> turn right <sep> go straight <sep> turn around </modified2> . <original3> walk down </original3> <modified3> stop in <sep> walk away from </modified3> the hallway and <original4> stop in </original4> <modified4> go into <sep> turn left at <sep> turn right at <sep> walk away from </modified4> the <original5> first </original5> <modified5> second <sep> third <sep> fourth <sep> last </modified5> doorway <original6> on your left </original6> <modified6> on your right <sep> straight ahead </modified6> .

\textbf{Input:} Exit the bathroom. Walk forward and go down the stairs. Stop four steps from the bottom.

\textbf{Output:}
<original1> exit </original1> <modified1> enter </modified1> the bathroom . <original2> walk forward </original2> <modified2> go backward <sep> turn left <sep> turn right </modified2> and <original3> go down </original3> <modified3> go up <sep> stop by <sep> walk away from </modified3> the stairs . stop <original4> four </original4> <modified4> one <sep> two <sep> three </modified4> steps from the <original5> bottom </original5> <modified5> top </modified5> .

\textbf{Input:} walk through open door, turn left, walk toward fireplace turn right, stop outside doorway.

\textbf{Output:}
<original1> walk through </original1> <modified1> walk past </modified1> open door , <original2> turn left </original2> <modified2> turn right <sep> turn around <sep> go straight </modified2> , <original3> walk toward </original3> <modified3> walk away from </modified3> fireplace <original4> turn right </original4> <modified4> turn left <sep> turn around <sep> go straight </modified4> , stop <original5> outside </original5> <modified5> inside </modified5> doorway .

    }%
}

\subsection{Model Variants}
\label{app:model_variants}

\paragraph{HEAR-SameEnvSwap.} 
This system is identical to \ourmodel, but the synthetic hallucinations are created using different strategies.
In the case of object hallucination, rather than swapping two objects within the same instruction, we replace an object in the instruction with another object randomly selected from those encountered along the described route. For room perturbation, instead of replacing a room mentioned in the instructions with another room from a list, we substitute it with another room that exists in the same environment.

\paragraph{One-stage \ourmodel.} This underlying model of this system is similar to the hallucination detection model of \ourmodel.
But its positive examples contain instructions with an empty token \texttt{[REMOVE]}. For example:

\textit{Positive: Go forward toward the windows. Exit \texttt{[BH]} \texttt{[REMOVE]} \texttt{[EH]} to living room.}

\textit{Negative: Go forward toward the windows. Exit \texttt{[BH]} exercise room \texttt{[EH]} to living room.}

Thus, instead of using two models as in \ourmodel, we can use this single model to score any correction, including deletion corrections. Concretely, with this model, we simply set the score function $R(\hat{\vec x}) = 1 - P(y = 1 \mid \hat{\vec x})$ where $P(y = 1 \mid \hat{\vec x})$ is the probability output by the model.
The training data of this model contain 216,323 pairs of positive and negative examples.

\subsection{Hyperparameters and Tools}
\label{app:models}

\begin{table}[t]
\centering
\footnotesize
\setlength{\tabcolsep}{3pt}
\begin{tabular}{@{}lc@{}}
\toprule
Hyperparameter                   & Value        \\ \midrule
Learning rate                   & $10^{-5}$         \\ 
Batch size                      & 128               \\
Optimizer                       & AdamW             \\
Training iterations     & $5 \times 10^{5}$ \\
Maximum instruction length         & 60                \\
Image feature size              & 2048              \\
Embedding dropout               & 0.1               \\
Hidden size                     & 768               \\
Transformer layers & 12                \\
Transformer dropout rate   & 0.1               \\
Number of parameters    & 250M              \\
Computation and training time   & RTX A4000: $\sim$72h    \\ \bottomrule
\end{tabular}
\caption{The hyperparameters of the hallucination detection and hallucination type classification models.}
\label{tab:hyperparams}
\end{table}

The hyperparameters and computation cost of the \ourmodel's two models are listed in \autoref{tab:hyperparams} (they have the same architecture and are trained in the same way). Other baseline models (\autoref{app:model_variants}) also have the same hyperparameters.
We implement our models with Pytorch 1.7.1, Huggingface Transformers 4.5.1, NLTK 3.6.7, and use SciPy 1.6.0 for our result analyses.

\subsection{Main Result Table}
\begin{table}[]
\centering
\begin{tabular}{@{}lccc@{}}
\toprule
System                 & Success Rate ↑ & Navigation Error ↓ & Checks  \\ \midrule
No communication       & 68.9 ± 7.1       & 6.6 ± 1.6            & 2.9 ± 0.6 \\
HEAR (no suggestion)   & 75.6 ± 6.6       & 4.7 ± 1.2            & 3.4 ± 0.7 \\
HEAR                   & 77.8 ± 6.3       & 4.6 ± 1.2            & 4.1 ± 0.8 \\
Oracle (no suggestion) & ~~~81.1 ± 6.0 $^{\dagger}$      & ~~3.4 ± 0.9  $^{\dagger}$          & 3.5 ± 0.7 \\
Oracle                 & ~~~87.8 ± 5.0 $^{\ddagger}$      & ~~2.7 ± 0.7 $^{\ddagger}$           & 3.6 ± 0.6 \\ \bottomrule
\end{tabular}
\caption{Performance measured by success rate (SR ↑) and navigation error (DIST ↓), and the number of check-button clicks recorded when human users perform navigation tasks with different assistant systems. 
The error bars after ± represent 85\% confidence intervals.
The symbols $^{\ddagger}$ and $^{\dagger}$ indicate results that are significantly higher than those of the ``No communication'' system in the first row, with $p < 0.004$ (Bonferroni correction) and $p < 0.05$, respectively, as determined by a two-related-sample t-test.}
\label{tab:main_result}
\end{table}

\autoref{tab:main_result} shows human navigation performance when using different assistant systems, which corresponds to the charts in \autoref{fig:human_sr}.

\begin{figure}[t!]
\centering
\includegraphics[width=0.5\textwidth]{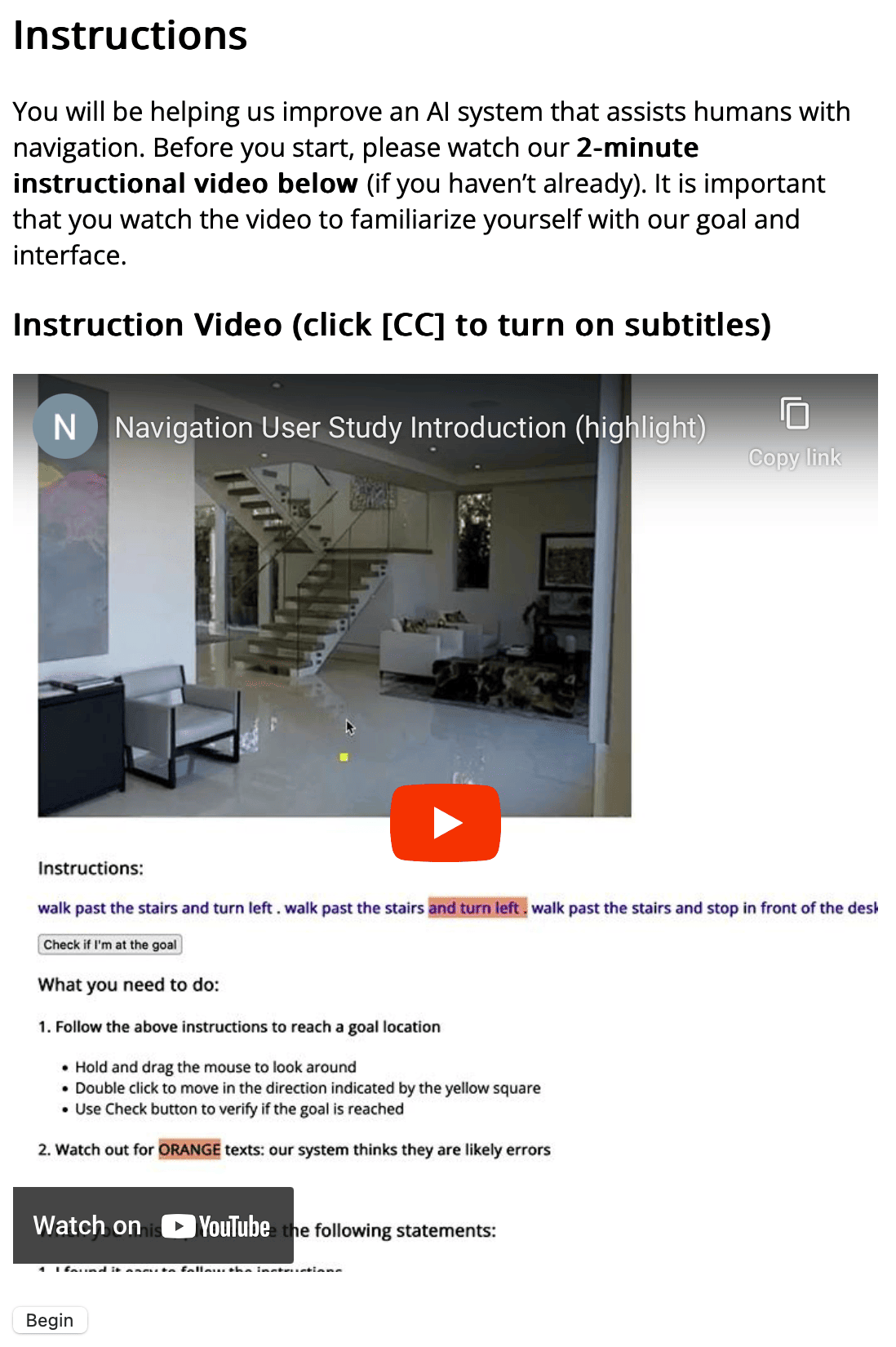}
\caption{Introductory page of the human navigation task. A video instruction is provided.}
\label{fig:human_eval_first}
\end{figure}

\begin{figure}[t!]
\centering
\includegraphics[width=0.5\textwidth]{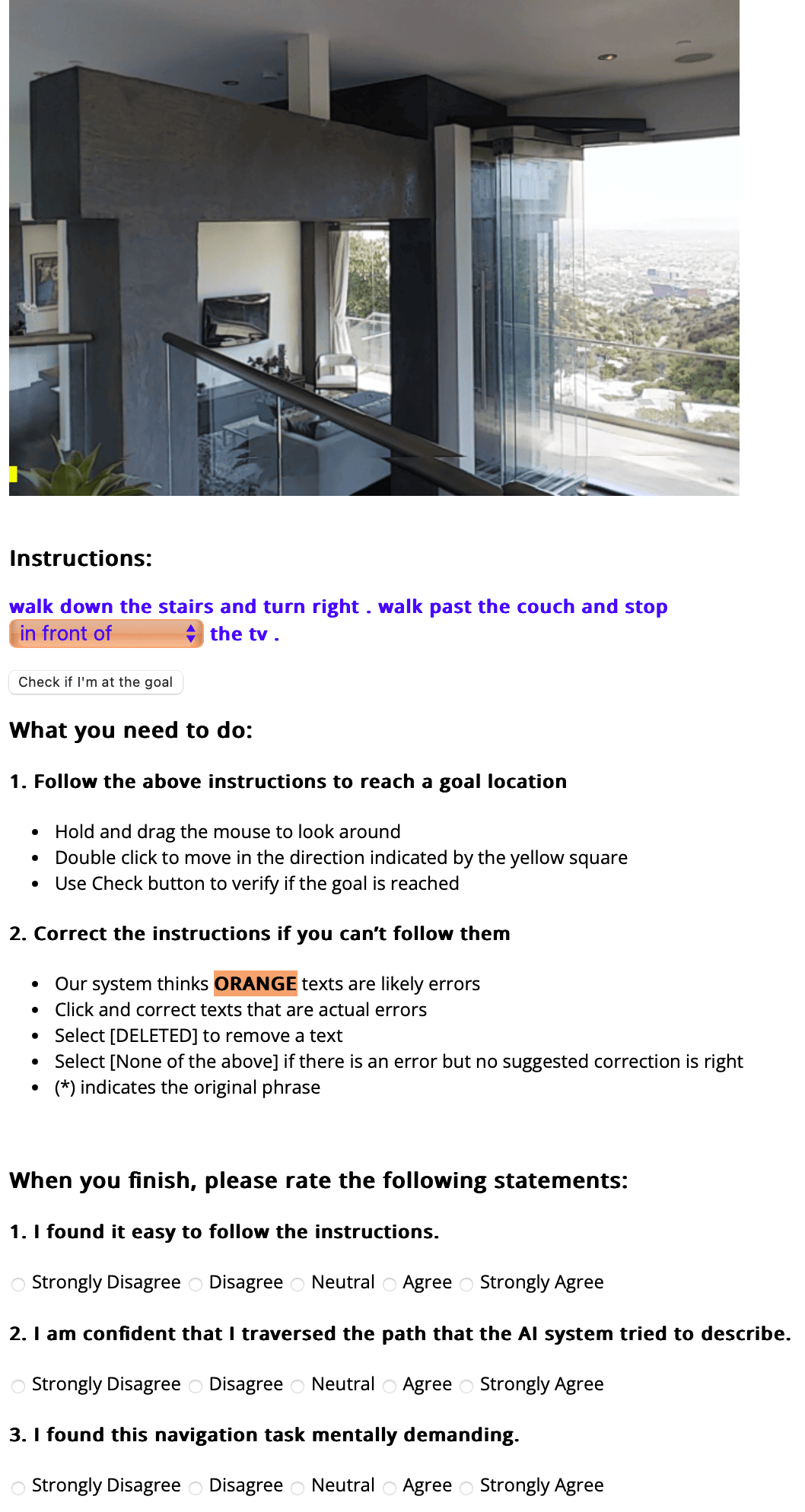}
\caption{The interface used by the \ourmodel and Oracle systems.}
\label{fig:eval_highlight_alternative}
\end{figure}

\begin{figure}[t!]
\centering
\includegraphics[width=0.5\textwidth]{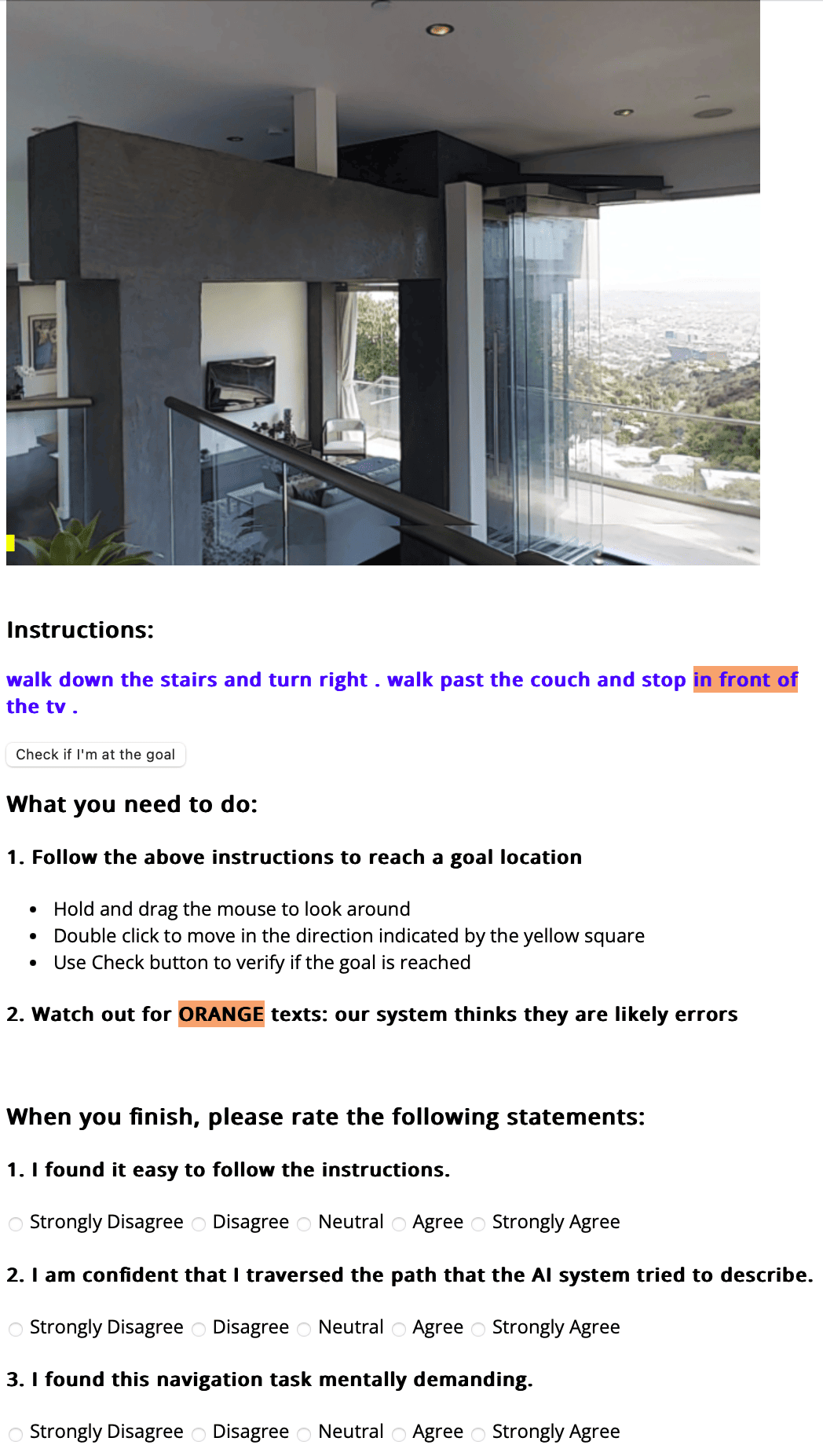}
\caption{The interface used by the \ourmodel and Oracle systems without correction suggestions.}
\label{fig:eval_highlight}
\end{figure}

\begin{figure}[t!]
\centering
\includegraphics[width=0.5\textwidth]{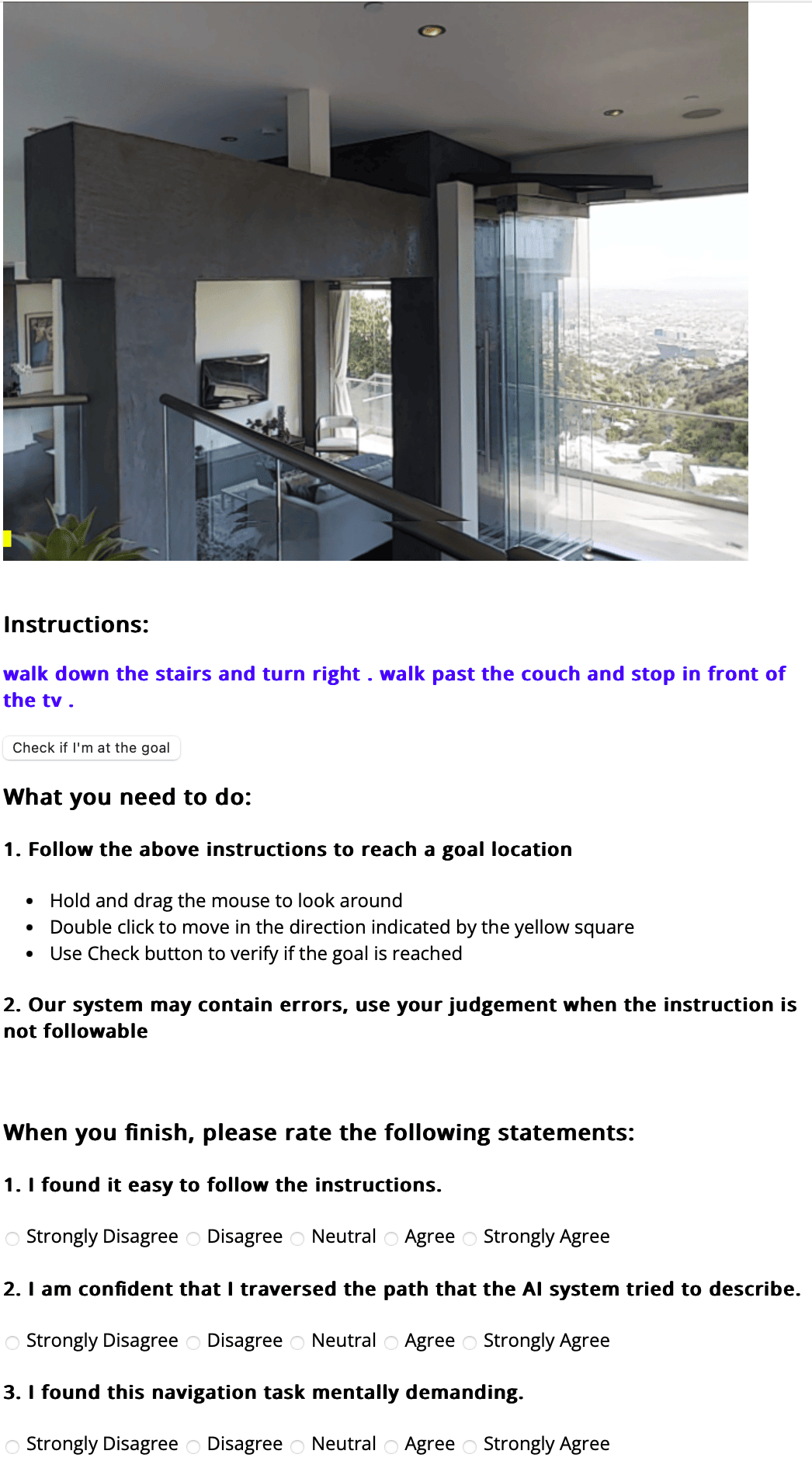}
\caption{The interface without highlights and suggestions (no communication).}
\label{fig:eval_no_communication}
\end{figure}

\subsection{Human Evaluation}
\label{app:human_evaluation}

\autoref{fig:eval_highlight_alternative} shows the user interface of the \textit{\ourmodel} and the \textit{Oracle} systems. 
\autoref{fig:eval_highlight} presents the interface of the \textit{\ourmodel (no suggestion)} and \textit{Oracle (no suggestion)} systems.  
\autoref{fig:eval_no_communication} is the interface of \textit{No communication}.
The interfaces are adapted from \citet{zhao2023cognitive} with the MIT License and Pangea\footnote{https://github.com/google-research/pangea} with the Apache License v2.0. 
Before starting a task, we provide the user with a video instruction that shows them how to use the interface (\autoref{fig:human_eval_first}). After they complete the task, we record their route, the number of times they click on the \textit{Check} button, and their subjective ratings.  
User participants must be at least 18 years old and speak English. 
The intended use of the system is first explained to them, and if they consent to perform the task, then they will be taken to the interface. 

This study has been approved by the Institutional Review Board (IRB). For data anonymization, we removed the only PII information, the Amazon Mechanical Turk ID, after collecting the data. 
This information will also be removed in the future dataset release and replaced with serial numbers that do not reveal the identities of the participants.
The dataset will be released under MIT license terms that are compatible with those of the tools used to create it and will be intended for research usage. 
We do not identify any potential risk to participants or the general public in releasing our dataset.

\begin{figure}[t]
\centering
\includegraphics[width=0.45\textwidth]{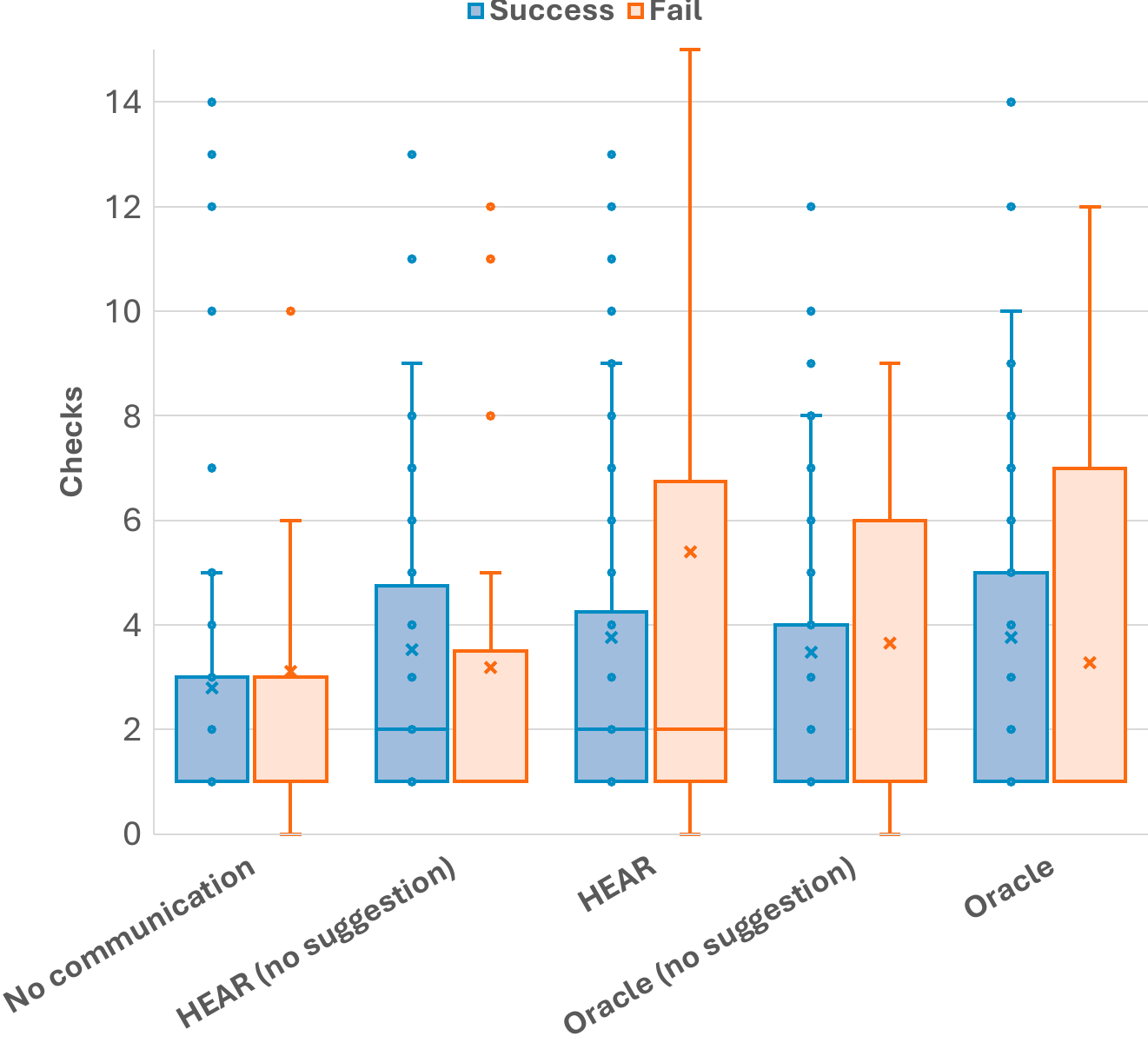}
    \caption{Number of check-button clicks when users succeed and fail on the task. \looseness=-1
    }
    \label{fig:checks_group}
\end{figure}

\subsection{Check Button Usage}

In \autoref{fig:checks_group}, we show the number of checks when users succeed or fail. 
We observe that highlights and suggestions increase the number of checks in both cases.

\begin{figure}[t]
\centering
\begin{subfigure}{\textwidth}
\includegraphics[width=0.6\textwidth]{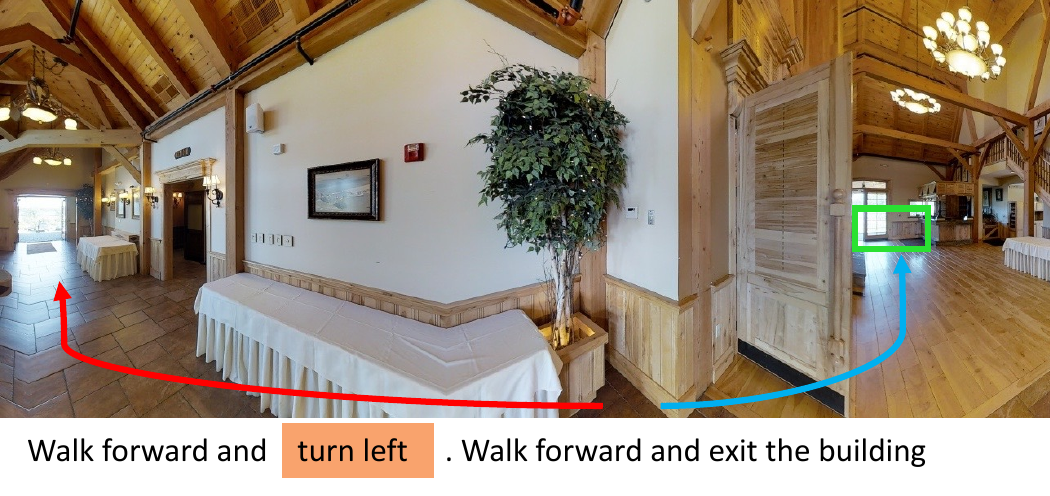}
\centering
    \caption{A qualitative example where our system accurately highlights a hallucinated direction and helps a user navigate successfully. Another user, who is not given the highlight, follows the instruction and takes the wrong turn. \looseness=-1
    }
\end{subfigure}\\
\begin{subfigure}{0.45\textwidth}
\includegraphics[width=0.9\textwidth]
{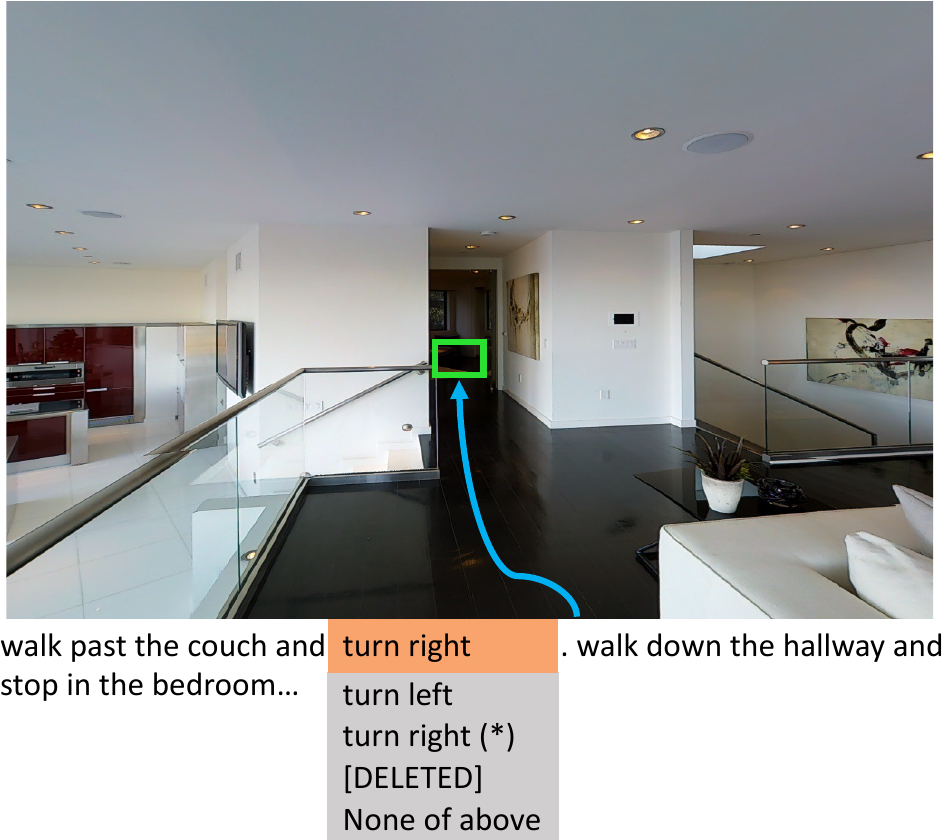}
\centering
\caption{Accurate highlights from our system help a user to correctly go straight. Although the suggestions are not accurate, it can still enable the user to make the right decision. \looseness=-1}
\end{subfigure}
\hfill
\begin{subfigure}{0.45\textwidth}
\includegraphics[width=0.9\textwidth]{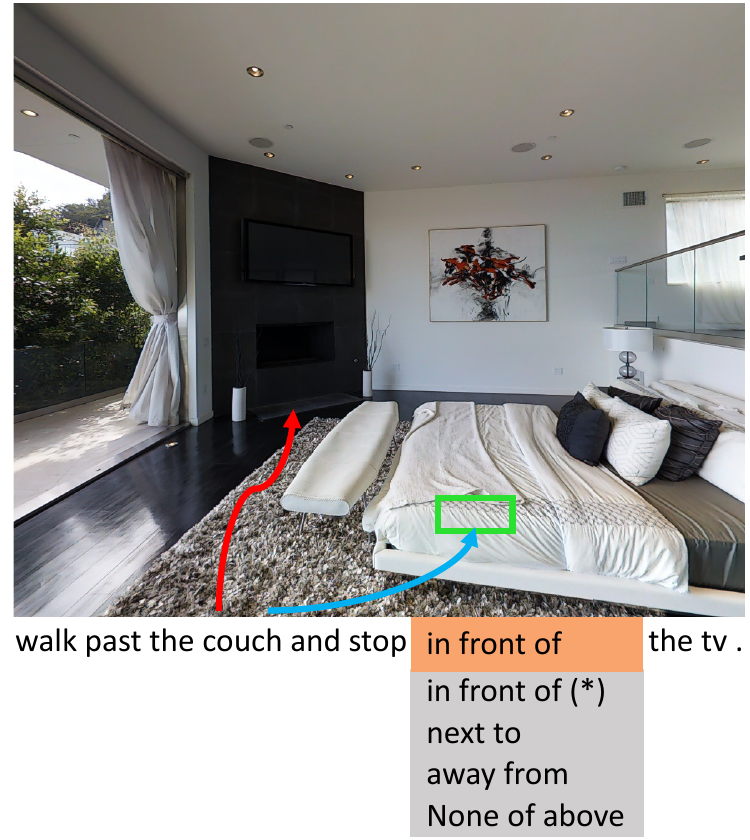}
\centering
\caption{In this case, the correct instruction is: walk past the couch and stop in front of the \textit{bed}. Inaccurate highlight generated by our system leads the user to the wrong location.  \looseness=-1}
\end{subfigure}
\caption{Additional qualitative examples. The true route and the target destination are marked by a \textcolor{blue}{blue} arrow and a \textcolor{green}{green} box, respectively. The user's route is indicated by a \textcolor{red}{red} arrow.}
\label{fig:more_qualtitative_examples}
\end{figure}

\subsection{Qualitative example (\autoref{fig:more_qualtitative_examples})}
\label{app:qualitative_examples}

